\def\year{2021}\relax
\documentclass[letterpaper]{article} 
\usepackage{aaai21}  
\usepackage{times}  
\usepackage{helvet} 
\usepackage{courier}  
\usepackage[hyphens]{url}  
\usepackage{graphicx} 
\usepackage{natbib}  
\usepackage{caption} 
\usepackage{amsmath}
\usepackage{amssymb}
\usepackage{multirow} 
\usepackage{booktabs}
\usepackage[colorlinks=true,citecolor=black]{hyperref}
\title{Non-Autoregressive Coarse-to-Fine Video Captioning}
\author{Paper ID 405}
\author {
        Bang Yang,\textsuperscript{\rm 1}
        Yuexian Zou, \textsuperscript{\rm 1,2}\thanks{Corresponding author.}
        Fenglin Liu, \textsuperscript{\rm 1} 
        Can Zhang \textsuperscript{\rm 1}\\
}
\affiliations {
    \textsuperscript{\rm 1} ADSPLAB, School of ECE, Peking University, Shen Zhen, China \\
    \textsuperscript{\rm 2} Peng Cheng Laboratory \\
    \{yb.ece, zouyx, fenglinliu98, zhangcan\}@pku.edu.cn
}

\begin{document}
\maketitle
\begin{abstract}
It is encouraged to see that progress has been made to bridge videos and natural language. However, mainstream video captioning methods suffer from slow inference speed due to the sequential manner of autoregressive decoding, and prefer generating generic descriptions due to the insufficient training of visual words (\textit{e.g.}, nouns and verbs) and inadequate decoding paradigm. In this paper, we propose a non-autoregressive decoding based model with a coarse-to-fine captioning procedure to alleviate these defects. In implementations, we employ a bi-directional self-attention based network as our language model for achieving inference speedup, based on which we decompose the captioning procedure into two stages, where the model has different focuses. Specifically, given that visual words determine the semantic correctness of captions, we design a mechanism of generating visual words to not only promote the training of scene-related words but also capture relevant details from videos to construct a \textit{coarse-grained} sentence ``template''. Thereafter, we devise dedicated decoding algorithms that fill in the ``template'' with suitable words and modify inappropriate phrasing via iterative refinement to obtain a \textit{fine-grained} description. Extensive experiments on two mainstream video captioning benchmarks, \textit{i.e.}, MSVD and MSR-VTT, demonstrate that our approach achieves state-of-the-art performance, generates diverse descriptions, and obtains high inference efficiency. Our code is available at \url{https://github.com/yangbang18/Non-Autoregressive-Video-Captioning}.
\end{abstract}

\section{Introduction}
Video captioning aims to automatically describe video contents with plausible sentences, which could be helpful for video retrieval, assisting visually-impaired people and so on. In recent years, neural captioning methods have risen to prominence and they generally adopt the encoder-decoder framework \cite{venugopalan2015translating}, where videos are encoded to sequences of vectors with Convolutional Neural Networks (CNNs), and captions are often decoded from these vectors via Recurrent Neural Networks (RNNs) or Transformer \cite{hori2017attention}. For real-time industrial applications, a good video captioning system may have low inference latency and describe scene-related details. However, existing methods mostly have deficiencies in both aspects.

\begin{figure}[t]
\centering
\includegraphics*[width = 0.99\linewidth]{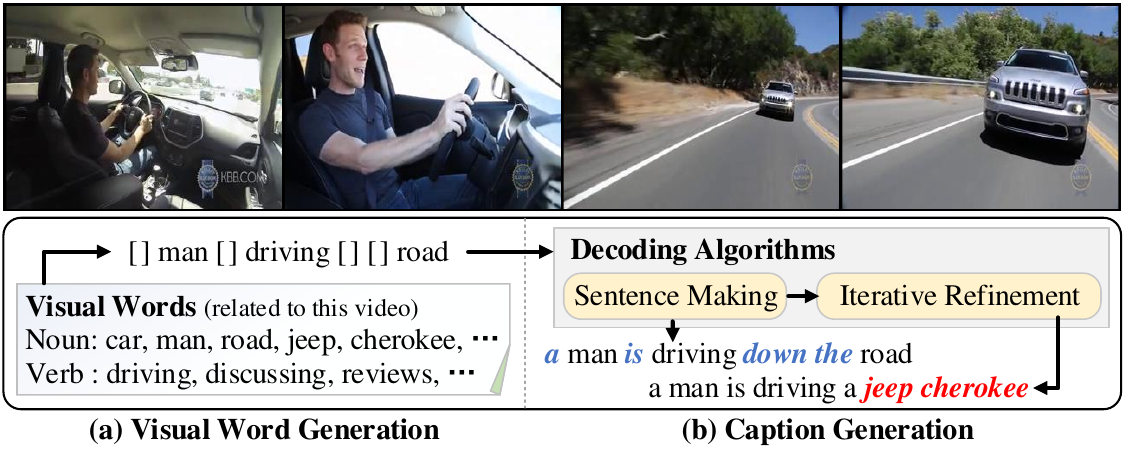}
\caption{Illustration of the proposed coarse-to-fine captioning procedure, where (a) visual words (\textit{i.e.}, nouns and verbs in this paper) are generated in parallel first to form a \textit{coarse-grained} ``template'', based on which (b) a \textit{fine-grained} description is yielded via dedicated decoding algorithms.}
\label{fig:ours}
\end{figure}

For caption generation, current methods are stick to \textit{autoregressive} (AR) decoding, \textit{i.e.}, conditioning each word on the previously generated outputs. Such sequential manner results in high inference latency, which is especially amplified by the fact that fine-grained descriptions are generally long \cite{gella2018story}.  Recently, \textit{non-autoregressive} (NA) decoding that generates words in parallel to achieve significant inference speedup becomes an emerging focus in neural machine translation (NMT) \cite{gu2018nonautoregressive, wang2019non, shao2020minimizing}. Nevertheless, NA decoding suffers from poor approximation to the target distribution after removing the sequential dependency \cite{gu2018nonautoregressive}, leading to a large performance gap compared with AR counterparts. 

For describing scene-related details, \textit{visual words} (\textit{e.g.}, nouns and verbs), which are visually-grounded and highly associated with semantic correctness \cite{song2017hierarchical}, deserve more attention than \textit{non-visual words} (\textit{e.g.} determiners and prepositions). By analyzing the existing caption corpora \cite{chen2011collecting,xu2016msr}, it is found that visual words are much less than non-visual words in number. Therefore, treating these two kinds of words equally to train a captioning model, as most of previous works do \cite{aafaq2019spatio, huang2020image}, will cause insufficient training of meaningful words, which could manifest as the lack of relevant details and diversity in generated descriptions \cite{dai2018neural,sammani2020show}. Besides, it is challenging for the models to generate satisfying captions with the proneness of error accumulation \cite{bengio2015scheduled}. So a flexible decoding paradigm that supports word modification is also needed.

In this paper, we propose a Non-Autoregressive Coarse-to-Fine (NACF) model to tackle the slow inference speed and unsatisfied caption quality concerns in video captioning. For achieving inference speedup, we employ a bi-directional self-attention based network \cite{vaswani2017attention} as our language model and train it with masked language modeling objective \cite{devlin-etal-2019-bert} so that any subset of target words can be predicted simultaneously based on the rest ones. For improving caption quality, we propose an alternative paradigm to decompose the captioning procedure into two stages, where the model has different focuses. Specifically, we propose a mechanism of generating visual words, \textit{i.e.}, \textit{visual word generation}, to not only promote the training of scene-related words but also require the model to generate a \textit{coarse-grained} sentence ``template'' at the first stage of inference phase. As shown in Fig. \ref{fig:ours} (a), the generated words in the ``template'' (\textit{e.g.}, ``driving'') summarize a first-glance gist of the scene, which could be instructive for the subsequent generation process. Thereafter, we devise dedicated decoding algorithms to produce \textit{fine-grained} descriptions at the second stage. As shown in Fig. \ref{fig:ours} (b), decoding algorithms first complete the ``template'' by filling in suitable words, and then if necessary, iteratively mask out and reconsider some inappropriate words that the language model is least confident about to ensure sentence fluency or capture more relevant details, \textit{e.g.}, a more precise phrase (``jeep cherokee'') is generated after deliberation. 

Our main contributions are summarized as follows:
\begin{itemize}
    \item We propose a Non-Autoregressive Coarse-to-Fine model to deal with slow inference speed and unsatisfied caption quality problems for video captioning.
    \item We design a mechanism of generating visual words and devise dedicated decoding algorithms to achieve coarse-to-fine rather than word-by-word captioning procedure for capturing more visually-grounded details from videos.
    \item Extensive experiments on MSVD and MSR-VTT demonstrate the effectiveness of our approach, which achieves state-of-the-art performance, generates diverse descriptions, and obtains high inference efficiency.
\end{itemize}

\section{Related Work}
\textbf{Video Captioning.} With the rapid development in deep learning, the neural captioning methods that follow the encoder-decoder framework have risen to prominence. One of the first works that adopt such framework is \cite{venugopalan2015translating}, where captions are generated by LSTM given the mean pooled representation over all frames. Later in \cite{yao2015describing,song2017hierarchical}, temporal attention is proposed to adaptively determine which subset of frames to focus on at each decoding step. In particular, \citeauthor{song2017hierarchical} propose to implicitly distinguish the importance of visual and non-visual words with a hierarchical attention mechanism \cite{song2017hierarchical}. Besides exploiting the temporal structure of videos, the utilization of multi-modalities and high-level semantics also draws great attention. For instance, \citeauthor{xu2017learning} propose multimodal attention to selectively focus on content-related modalities \cite{xu2017learning}. \citeauthor{liu2019aligning} leverage a pre-trained network to extract attributes, which are defined as the properties observed in visual contents with rich semantic cues, and use them to align visual features \cite{liu2019aligning}. Most recently, \citeauthor{huang2020image} propose to couple attribute prediction with caption generation in an end-to-end manner \cite{huang2020image}. \citeauthor{pan2020spatio} propose to exploit spatio-temporal object interactions and distill such knowledge into the captioning model \cite{pan2020spatio}. However, all these methods adopt sequential decoding and treat visual and non-visual words equally in terms of the loss function.

\textbf{Non-Autoregressive Decoding.} Due to high inference efficiency, NA decoding has aroused widespread attention in the community of NMT. By removing the sequential dependency, NA decoding can generate all words in one shot to speed up decoding \cite{gu2018nonautoregressive} but at the cost of inferior accuracy that manifests as token repetitions in the generated outputs \cite{wang2019non,guo2019non,lee-etal-2018-deterministic}. To compensate for the performance degradation, \citeauthor{guo2019non} integrate strong conditional signals into the decoder inputs to benefit the learning of internal dependencies within a sentence \cite{guo2019non}. Besides one-shot generation, some works propose to iteratively refine the sentences, so that the model can condition on parts or the whole of the previous outputs \cite{lee-etal-2018-deterministic,ghazvininejad2019constant,gu2019levenshtein,mansimov2019generalized}. But the downside of these methods is that from-scratch parallel generation, \textit{i.e.}, employing the completely unknown sequences as the decoder inputs, often leads to translation errors in the early stages due to insufficient context, which could greatly influence the subsequent predictions.

\textbf{Summary.} Rather than designing a sophisticated architecture, our work contributes by proposing an alternative decoding paradigm, \textit{i.e.}, generating descriptions from coarse-grained to fine-grained with the scheme of NA decoding, to generate semantically correct video captions with higher efficiency. This work pursues the iterative approaches in NMT \cite{ghazvininejad2019constant}, but the core difference is that we propose to capture visual words first to formulate partially observed sequences, which can provide rich contextual information with the model to alleviate description ambiguity and thus eventually enhance the caption quality.

\section{Approach}
In this section, we first introduce the architecture of our Non-Autoregressive Coarse-to-Fine (NACF) model, followed by the proposed visual word generation. Then, we describe the coarse-to-fine captioning procedure during inference, where three dedicated decoding algorithms are presented. 

\begin{figure}[t]
\centering
\includegraphics*[width = 0.99\linewidth]{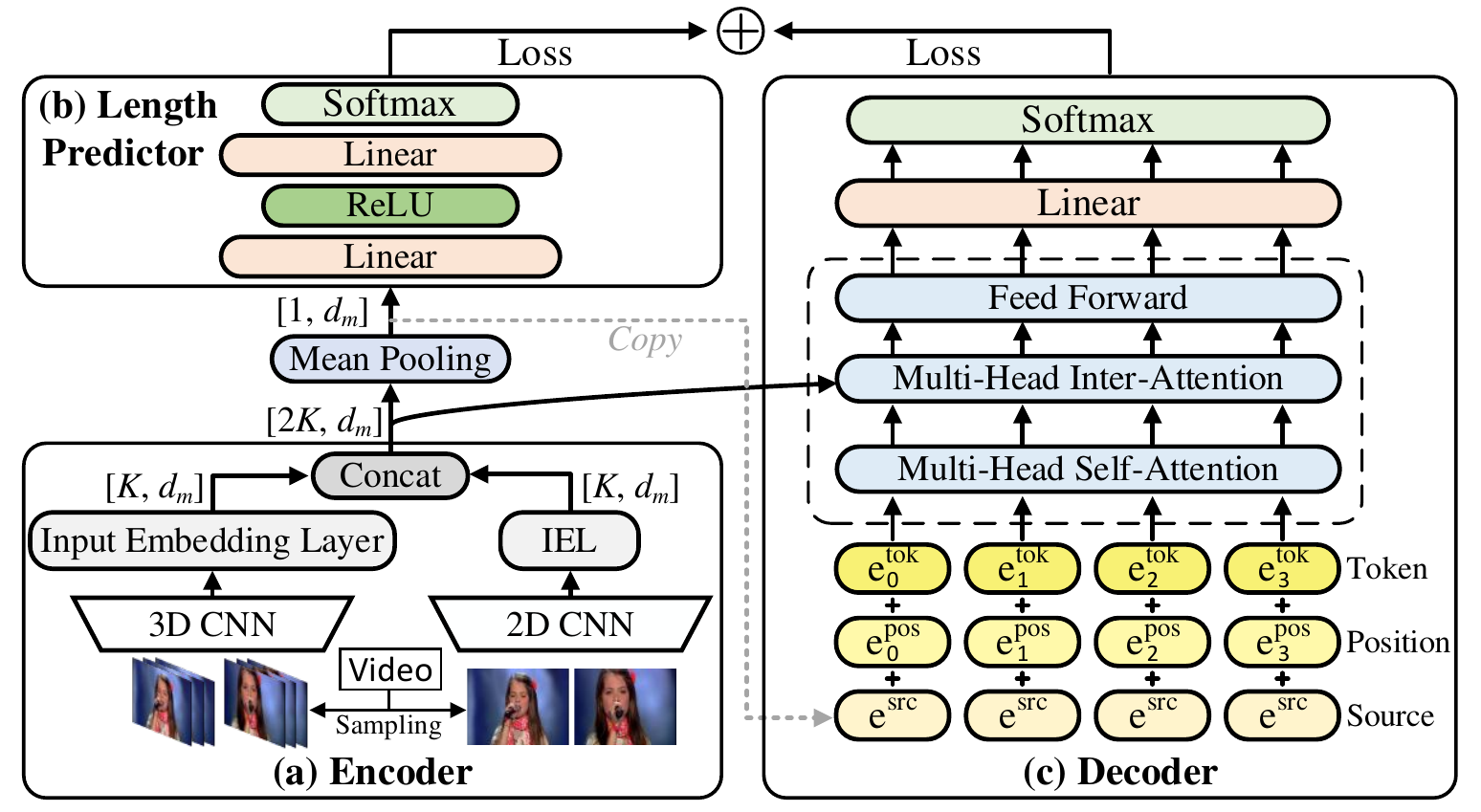}
\caption{An overview of our proposed NACF architecture, which comprises a CNN-based encoder, a length predictor module and a bi-directional self-attention based decoder.} 
\label{fig:framework}
\end{figure}

\subsection{Architecture}\label{sec:architecture}
As shown in Fig. \ref{fig:framework}, our NACF comprises three modules: a CNN-based encoder, a length predictor and a bi-directional self-attention based decoder.

\textbf{Encoder.} Given a sequence of video frames/clips of length $K$, we feed it into pre-trained 2D/3D CNNs to obtain visual features $V = \{v_k\}^K_{k=1} \in \mathbb{R}^{K\times d_v}$, which are further encoded to compact representations $R \in \mathbb{R}^{K\times d_m}$ via a input embedding layer (IEL), \textit{i.e.}, $R = f_{IEL}(V)$. Here $f_{IEL}$ adopts the shortcut connection in highway networks \cite{srivastava2015training}, so it can be formalized as follows (omitting biases for clarity):
\begin{equation}
\begin{split}
    f_{IEL}(V) &= {\rm BN}(G \circ \overline{V} + (1-G) \circ \hat{V})\\
    \overline{V} &= VW_{e1}\\
    \hat{V} &= \tanh{(\overline{V}W_{e2})}\\
    G &= \sigma(\overline{V}W_{e3})
\end{split}
\end{equation}
where ${\rm BN}$ denotes batch normalization \cite{ioffe2015batch}, $\circ$ is the element-wise product, $\sigma$ is sigmoid function, $W_{e1} \in \mathbb{R}^{d_v\times d_m}$, and $\{W_{e2},W_{e3}\} \in \mathbb{R}^{d_m\times d_m}$. When considering multi-modalities, \textit{e.g.}, image and motion modalities, we simply apply concatenation to obtain $R \in \mathbb{R}^{2K\times d_m}$.

\textbf{Length Predictor.} Unlike AR decoding that can automatically decide the sequence length $N$ by predicting the end-of-sentence token, NA decoding must know $N$ ahead. So a length predictor (LP) is introduced to predict length distribution $L \in \mathbb{R}^{N_{max}}$ given the outputs $R$ of the encoder:
\begin{equation}
\label{eq:lp}
L = f_{LP}(R) = {\rm Softmax}({\rm ReLU}({\rm MP}(R)W_{l1})W_{l2})
\end{equation}
where ${\rm MP}$ denotes mean pooling, $W_{l1} \in \mathbb{R}^{d_m\times d_m}$, $W_{l2} \in \mathbb{R}^{d_m \times N_{max}}$, and $N_{max}$ is the predefined maximum sequence length. Given the ground-truth length distribution $L^*$, whose $j$-th element denotes the percentage of sentences of length $j$ in the training corpus for a specific video, we minimize the Kullback-Leibler (KL) divergence between $L$ and $L^*$:
\begin{equation}
\label{eq:length_loss}
\mathcal{L}_{len} = D_{KL}(L^* || L) = -\sum_{j=1}^{N_{max}}{l^*_j \log \frac{l_j}{l^*_j}}
\end{equation} 
During training, we directly use the sequence length of ground-truth sentences. As for inference, we will depict the utilization of the predicted length distribution $L$ in experimental settings.

 
\textbf{Decoder.} To obtain a non-autoregressive decoder, we adopt one-layer decoder of Transformer \cite{vaswani2017attention} with two modifications. One is that we remove the causal mask in the self-attention layer. By doing so, our decoder becomes bi-directional, thus the prediction of each token can use both left and right contexts. Another is that we pursue the works in NMT \cite{lee-etal-2018-deterministic,guo2019non} to enhance decoder inputs by integrating the copied source information (the dashed line in Fig. \ref{fig:framework}(c)). To train the model, we use the masked language modeling objective (\textit{a.k.a.} the cloze task) in BERT \cite{devlin-etal-2019-bert}. Specifically, some tokens in a ground-truth sentence $Y^*$ are randomly masked out to obtain a partially-observed sequence $Y_{obs}$ and a masked (unobserved) sequence $Y_{mask} = Y^* \setminus Y_{obs}$. Then the decoder takes $Y_{obs}$ and representations $R$ as inputs to predict the probability distribution over words:
\begin{equation}
\label{decoder}
    p_{\theta}(y | Y_{obs}, R) = f_{dec}(Y_{obs}, R)
\end{equation}
where $f_{dec}$ denotes the transformation within the decoder\footnote{Detailed formulation of $f_{dec}$ is left to the technical appendix.}. We only minimize the negative log likelihood of $Y_{mask}$:
\begin{equation}
\label{eq:mask_loss}
\mathcal{L}_{mlm} = - \sum_{y \in Y_{mask}} \log p_{\theta}(y | Y_{obs}, R)
\end{equation}
Unlike BERT that uses a small masking ratio (\textit{e.g.}, 15\%), we use a uniformly distributed ratio ranging from $\beta_{l}$ to $\beta_{h}$ so that the model can be trained with examples of different difficulties. Next, we will elaborate on how to generate meaningful visual words with the proposed decoder.

\subsection{Visual Word Generation}\label{sec:visual_word_generation}
For non-autoregressive video captioning, visual word generation is proposed for two purposes: promoting the training of scene-related words and generating coarse-grained ``templates'' that serve as starting points at the inference phase. To achieve that, we directly use the proposed decoder without introducing extra parameters. Formally, given a ground-truth sentence $Y^*$ of length $N$, we first construct a corresponding target sequence $Y^{vis} = \{y^{vis}_n\}^N_{n=1}$ as follows:
\begin{equation}
\label{eq:construct_vis_target}
    y^{vis}_n =
    \left\{\begin{array}{ll}{y^{*}_n} & {\text { if } {\rm POS}(y^*_n) \in \{{\rm noun, verb}\}} \\ {[mask]} & {\text { otherwise }}\end{array}\right.
\end{equation}
where ${\rm POS}(\cdot)$ denotes the part-of-speech of a word. Then the decoder is forced to predict $Y^{vis}$ without available word information, \textit{i.e.}, $Y_{obs}^{vis} = \varnothing_{[vis]}$ (a sequence of the same special token $[vis]$ in practice). Hence the loss function for visual word generation is defined as:
\begin{equation}
\label{eq:visual_loss}
\mathcal{L}_{vis} = - \sum_{y \in Y^{vis}} \log p_{\theta}(y |\varnothing_{[vis]}, R)
\end{equation}
As the generation process solely depends on $R$, the generated visual words may not be comprehensive. But as we will verify later, they are instructive for the follow-up caption generation process. Finally, the overall loss function of our approach is formulated as:
\begin{equation}
\label{eq:overall_loss}
\mathcal{L}_{\rm NACF} = \mathcal{L}_{len} + \mathcal{L}_{mlm} + \lambda\mathcal{L}_{vis}
\end{equation}
where $\lambda$ is set to 0.8 empirically.

\subsection{Coarse-to-Fine Captioning}\label{sec:coarse_to_fine}
To yield plausible descriptions during inference, our captioning procedure is decomposed into two stages. At the first stage, we generate a coarse-grained ``template'' $Y^{(0)}$ and collect its confidence $C^{(0)}$ given the predicted sequence length $N$ and representations $R$: 
\begin{equation}
\label{eq:attribute}
    y^{(0)}_n, c^{(0)}_n = (\arg)\max_{w} p_\theta(y = w |\varnothing_{[vis]}, R)
\end{equation}
where $y^{(0)}_n$ is either the $[mask]$ token or a visual word (see Eq. \ref{eq:construct_vis_target}). In the subsequent generation process, four variables are introduced: the number of iterations $T$, the observed sequence at $t$-th ($t \in [1, T]$) iteration $Y^{(t)}_{obs}$, the prediction result $Y^{(t)}$ and its confidence $C^{(t)}$. Specifically, $Y^{(1)}_{obs}$ is initialized by the visual words in the coarse-grained ``template'' $Y^{(0)}$:
\begin{equation}
\label{eq:initialize}
    Y^{(1)}_{obs} = \{y^{(0)}_n|y^{(0)}_n\neq[mask] \}
\end{equation}
Then at the second stage, three dedicated decoding algorithms, \textit{i.e.}, Mask-Predict (MP) \cite{ghazvininejad2019constant}, Easy-First (EF) and Left-to-Right (L2R) are introduced to generate fine-grained descriptions. 

\begin{table*}[t]
    \centering
    \begin{tabular}{cccc|ccc}  
    \toprule 
    \multirow{2}{*}{Model} &\multicolumn{3}{c}{MSVD} &\multicolumn{3}{c}{MSR-VTT}\cr \cmidrule{2-7}
    &BLEU@4 &METEOR &CIDEr-D &BLEU@4 &METEOR &CIDEr-D\cr
    \midrule 
    STAT \cite{yan2019stat}
    &52.0 &33.3 &73.8 &39.3 &27.1 &43.8\cr
    
    GRU-EVE \cite{aafaq2019spatio} 
    &47.9 &35.0 &78.1 &38.3 &28.4 &48.1\cr
    
    POS-CG \cite{wang2019controllable} 
    &52.5 &34.1 &88.7 &\textbf{42.0} &28.2 &48.7\cr

    MARN \cite{pei2019memory} 
    &48.6 &35.1 &92.2 &40.4 &28.1 &47.1\cr
    
    MAD-SAP \cite{huang2020image}
    &53.3 &35.4 &90.8 &41.3 &28.3 &48.5\cr
    
    STG-KD \cite{pan2020spatio}
    &52.2 &\textbf{36.9} &93.0 &40.5 &28.3 &47.1\cr
    \midrule
    AR-B 
    &48.7 &35.3 &91.8 &40.5 &\textbf{28.7} &49.1\cr
    
    NA-B
    &53.7 &35.5 &92.8 &40.4 &28.0 &47.6\cr
    
    \textbf{Our NACF}
    &\textbf{55.6} &36.2 &\textbf{96.3} &\textbf{42.0}  &\textbf{28.7} &\textbf{51.4}\cr
    \bottomrule
    \end{tabular}
    \caption{Comparison with the state-of-the-art methods on MSVD and MSR-VTT, where AR-B and NA-B are our baselines.}
    \label{tab:MSVD_MSRVTT} 
\end{table*}

\textbf{Mask-Predict (MP).} This algorithm iterates over two steps at $t$-th iteration: \textit{Mask} where $m_t$ tokens with lowest confidence are masked out and \textit{Predict} where those masked tokens are reconsidered based on the rest $N-m_t$ tokens. The order of these two steps are switched in this paper because $Y^{(1)}_{obs}$ is already obtained. We only update the prediction results $Y^{(t)}$ and confidence $C^{(t)}$ for those unobserved tokens given $Y^{(t)}_{obs}$ and $R$:
\begin{equation}
\label{eq:predict}
    y^{(t)}_n, c^{(t)}_n = \left\{\begin{array}{ll}{(\arg)\max_{w} p_\theta(y = w |Y^{(t)}_{obs}, R)} & {\text { if }n \in I_t} \\ {y^{(t-1)}_n, c^{(t-1)}_n} & {\text { otherwise }}\end{array}\right.
\end{equation}
where $I_t$ denotes the index set of unobserved tokens:
\begin{equation}
\label{eq:I_t}
    I_t = \{n | y^{(t-1)}_n \notin Y^{(t)}_{obs}\}
\end{equation}
As low confidence $c^{(t)}_n$ means the token $y^{(t)}_n$ is incompatible with others, reconsidering such token could benefit caption quality. So for the next iteration, $Y^{(t+1)}_{obs}$ is defined as:
\begin{equation}
\label{eq:y_obs}
    Y^{(t+1)}_{obs} = \{y^{(t)}_j|j\in\mathop{{\rm topk}}\limits_n(C^{(t)}, k = N-m_{t+1})\}
\end{equation}
where we use a linear decay ratio $r$ to decide $m_t$ and make sure there is at least one token to be reconsidered:
\begin{equation}
    \label{eq:m_t}
    r = \frac{T-t+1}{T},\quad m_t = \max(\lfloor N\cdot r\rfloor, 1)
\end{equation}


\textbf{Easy-First (EF).} This algorithm generates $q$ tokens with highest confidence among the unobserved tokens at each iteration. Given $N$ and $u$ (the cardinality of $Y^{(1)}_{obs}$), EF algorithm needs $T = \lceil (N-u)/ q \rceil$ iterations and can be briefly formulated as: 
\begin{equation}
\label{eq:y_obs_ef}
    Y^{(t+1)}_{obs} = Y^{(t)}_{obs} \cup \{y^{(t)}_{I_t,j}|j\in\mathop{{\rm topk}}\limits_n(C^{(t)}_{I_t}, k = q)\}
\end{equation}
where $Y^{(t)}_{I_t}$ and $C^{(t)}_{I_t}$ denote the prediction and confidence of unobserved tokens at $t$-th iteration respectively. $Y^{(t)}$ and $C^{(t)}$ are calculated by Eq. \ref{eq:predict} while $I_t$ is computed by Eq. \ref{eq:I_t}. Since the visual words in $Y^{(1)}_{obs}$ are not modified during generation, we can reconsider them based on $Y=Y^{(T)}\setminus Y^{(1)}_{obs}$ and $R$: 
\begin{equation}
\label{eq:one_more_iteration}
    y^{(T)}_n, c^{(T)}_n = (\arg)\max_{w} p_\theta(y = w |Y, R)\quad\mathrm{s.t.}\ y^{(0)}_n\neq[mask]
\end{equation}

\begin{figure}[t]
\centering
\includegraphics*[width = 0.99\linewidth]{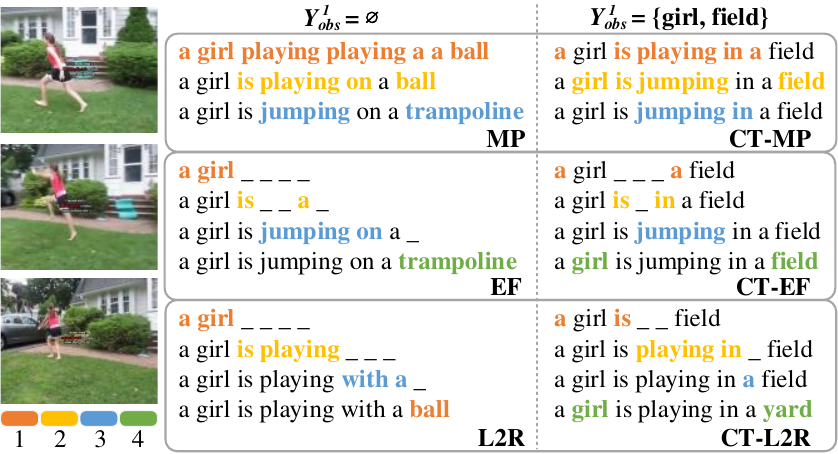}
\caption{Illustration of how to generate captions using different decoding algorithms (best viewed in color). The newly generated words (in color) are predicted based on the observed words (in black). We set $T=3$ for MP algorithm, $q=2$ for L2R and EF algorithms. The prefix ``CT-'' mean using the coarse-grained ``templates'' (Eq. \ref{eq:initialize}).} 
\label{fig:inference_example}
\end{figure}

\textbf{Left-to-Right (L2R).} In contrast to EF, this algorithm is monotonous, \textit{i.e.}, it generates $q$ tokens among the unobserved tokens from left to right at each iteration. L2R algorithm also needs $T = \lceil (N-u)/ q \rceil$ iterations and can be briefly defined as:
\begin{equation}
\label{eq:y_obs_l2r}
    Y^{(t+1)}_{obs} = Y^{(t)}_{obs} \cup \{y^{(t)}_{I_t, 1}, \dots, y^{(t)}_{I_t, q}\}
\end{equation}
where $Y^{(t)}_{I_t}$ denote the prediction of unobserved tokens at $t$-th iteration. $Y^{(t)}$ (also $C^{(t)}$) is computed by Eq. \ref{eq:predict} while $I_t$ is calculated by Eq. \ref{eq:I_t}. Similar to EF, one more iteration can be added to reconsider the visual words (Eq. \ref{eq:one_more_iteration}). Although we make both L2R and EF to generate $q$ words at each iteration, a fixed number of iterations $T$ can be set for both of them to produce captions in constant time. Besides, both L2R and EF algorithms can cooperate with the MP algorithm to iteratively refine sentences if necessary.


\textbf{Example.} Each of the decoding algorithms mentioned above can start with either the generated coarse-grained ``template'' (Eq. \ref{eq:initialize}) or a completed unknown sequence by setting $Y^{(1)}_{obs} = \varnothing$ and $I_1 = \{1, 2, \dots, N\}$ (Eq. \ref{eq:I_t}). To differentiate these two versions, we name the algorithms utilizing coarse-grained ``templates'' with the prefix ``CT-''. As shown in Fig. \ref{fig:inference_example}, the original version of algorithms, \textit{i.e.}, MP, EF and L2R, hallucinate some concepts (\textit{e.g.}, ``trampoline'' and ``ball'') that does not exist in visual contents due to the limited contextual information of completed unknown sequences. But under the guidance of some generated visual words (\textit{i.e.}, ``girl'' and ``field''), the algorithms can produce more precise descriptions, which will be verified later.

\section{Experiments}
In this section, we evaluate our NACF on two datasets: Microsoft Video Description (MSVD) \cite{chen2011collecting} and MSR-Video To Text (MSR-VTT) \cite{xu2016msr}. 

\subsection{Experimental Settings} \label{sec:implementation_details}
\textbf{Datasets.} MSVD contains 1,970 video clips and roughly 80,000 English sentences. We follow the split settings in prior works \cite{pei2019memory,pan2020spatio}, \textit{i.e.}, 1,200, 100 and 670 videos for training, validation and testing, respectively. MSR-VTT consists of 10,000 video clips, each of which has 20 captions and a category tag. Following the official split, we use 6,513, 497 and 2,990 videos for training, validation and testing, respectively. The vocabulary size of MSVD is 9,468, whereas that of MSR-VTT is 10,547.

\textbf{Feature Extraction.} We follow \cite{pei2019memory} and opt for the same type of features, \textit{i.e.}, 2048-D image features from ResNet-101 \cite{he2016deep} pre-trained on the ImageNet dataset \cite{deng2009imagenet}, 2048-D motion features from ResNeXt-101 with 3D convolutions \cite{hara2018can} pre-trained on the Kinetics dataset \cite{kay2017kinetics}, and all category tags included in MSR-VTT.

\textbf{Length Beam and Teacher Rescoring.} Following the common practice of noisy parallel decoding during inference \cite{gu2018nonautoregressive,wang2019non}, we select top $B$ length candidates from the predicted length distribution $L$, and decode the same example with different lengths in parallel. An autoregressive counterpart (\textit{i.e.}, the AR-B introduced later) is then used to re-score these $B$ candidates. We finally select captions with the highest confidence as hypotheses. 

\textbf{Parameter Settings.} The maximum sequence length $N_{max}$ is set to 20 for MSVD, whereas $N_{max}=30$ for MSR-VTT. We empirically set $K=8$ for each modality. For the decoder, we adopt 1 decoder layer, 512 model dimensions, 2,048 hidden dimensions and 8 attention heads per layer. Both word and position embeddings are implemented by trainable 512-D embedding layers. For regularization, we use 0.5 dropout and $5 \times 10^{-4}$ $L_2$ weight decay. We train batches of 64 video-sentence pairs using ADAM \cite{kingma2014adam} with an initial learning rate of $5 \times 10^{-3}$. We stop training our model until 50 epochs are reached. We use NLTK toolkit \cite{bird2009natural} for part-of-speech tagging. In the following experiments, our NACF uses CT-MP algorithm with the number of iterations $T$ of 5 and beam size $B$ of 6 unless otherwise specified.

\textbf{Evaluation Metrics.} We report three common metrics including BLEU \cite{papineni2002bleu}, METEOR \cite{banerjee2005meteor} and CIDEr-D \cite{vedantam2015cider}. All metrics are computed by Microsoft COCO Evaluation Server \cite{chen2015microsoft}. 

\textbf{Compared Approaches.} We compare our NACF with the following state-of-the-art methods: \textbf{STAT} \cite{yan2019stat} and \textbf{GRU-EVE} \cite{aafaq2019spatio} which capture spatio-temporal dynamics by attention mechanism and Short Fourier Transform respectively, \textbf{POS-CG} \cite{wang2019controllable} which uses the global syntactic part-of-speech information, \textbf{MARN} \cite{pei2019memory} which leverages memory network to capture cross-video contents, \textbf{MAD-SAP} \cite{huang2020image} which predicts a concise set of attributes at each decoding step, and \textbf{STG-KD} \cite{pan2020spatio} which distills the knowledge of spatio-temporal object interactions. Additionally, we consider two baselines, namely autoregressive baseline (\textbf{AR-B}) and non-autoregressive baseline (\textbf{NA-B}). Specifically, AR-B has similar architecture as our NACF, but it excludes the length predictor and includes causal mask in the self-attention layer. NA-B is the same with our NACF but it excludes visual word generation, \textit{i.e.},  $\lambda = 0$ (Eq. \ref{eq:overall_loss}).

\subsection{Performance Comparison}
The quantitative results in Table \ref{tab:MSVD_MSRVTT} illustrate that our NACF achieves state-of-the-art performance on both MSVD and MSR-VTT datasets, which is mainly beneficial from the proposed coarse-to-fine captioning procedure. Since we opt for the same type of features as MARN and similar features as MAD-SAP and STG-KD, fair comparisons between these methods and our approach can be guaranteed. Notably, our NACF achieves significant improvement on CIDEr-D, \textit{e.g.}, a relative improvement of 3.5\% on MSVD compared with STG-KD while 6.0\% on MSR-VTT compared with MAD-SAP. As the CIDEr-D metric is to punish the often-seen but uninformative n-grams in the dataset, the superior performance on CIDEr-D indicates that our NACF can capture more scene-related keywords from videos. It is noteworthy that our NACF is slightly worse than STG-KD on the METEOR metric in MSVD, which is because the latter learn spatio-temporal object interactions well in a small dataset with few portions of animations \cite{pan2020spatio}.

\begin{table}[t]
    \centering
    \begin{tabular}{c|l|ccc}  
    \toprule
    exp &Model &B4 &M &C$_D$\cr 
    \midrule
    1 & NA-B &40.4 &28.0 &47.6\cr
    2 & NA-B w/ $\mathcal{L}_{vis}$ &40.8 &28.2 &49.4\cr
    3 & NA-B w/ $\mathcal{L}_{vis}$, CT (\textbf{NACF}) &\textbf{42.0} &\textbf{28.7} &\textbf{51.4}\cr
    \midrule
    4 & AR-B &40.5 &28.7 &49.1\cr
    5 & AR-B w/ $\mathcal{L}_{vis}$ &\textbf{41.4} &\textbf{29.0} &\textbf{50.8}\cr
    \bottomrule
    \end{tabular}
    
    \caption{Effect of the visual word generation loss ($\mathcal{L}_{vis}$, Eq. \ref{eq:visual_loss}) and the generated coarse-grained templates (CT) on the performance on MSR-VTT in terms of BLEU@4 (B4), METEOR (M) and CIDEr-D (C$_D$).}
    \label{tab:different_components}  
\end{table}

\begin{figure}[t]
\centering
\includegraphics*[width = 0.99\linewidth]{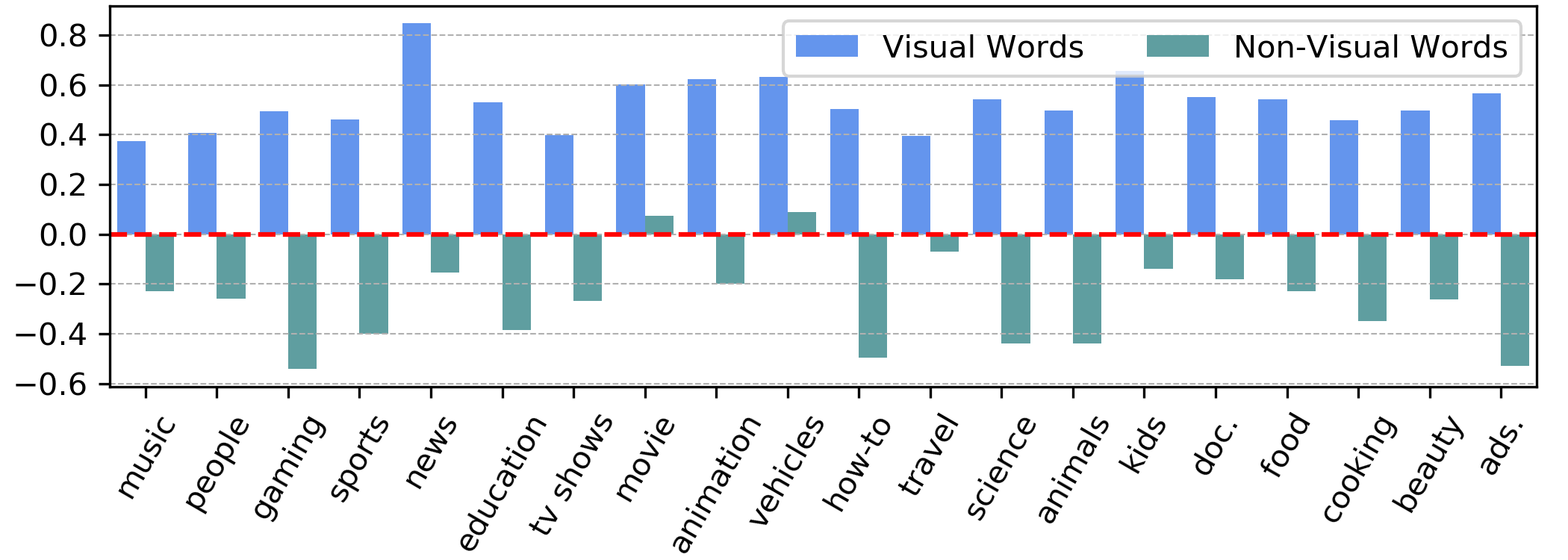}
\caption{The average relative growth rate of word frequency after training with visual word generation ($\mathcal{L}_{vis}$). Here we compare exp2 with exp1 (listed in Table \ref{tab:different_components}), and detail the results in all 20 categories of videos from MSR-VTT. ``doc.'' and ``ads.'' are short for documentary and advertisement.}
\label{fig:average_rgr}
\end{figure}

We also present the performance of baselines in Table \ref{tab:MSVD_MSRVTT}, and obtain two observations. (1) Compared with AR-B, NA-B obtains superior performance on MSVD while performs poorly on MSR-VTT, showing that the removal of sequential dependency makes NA decoding face more a severe ``multi-modality'' problem\footnote{For example, a NA model considers two possible captions $C_1$ and $C_2$, it could predict one token from $C_1$ while another token from $C_2$ due to the conditional independence.} \cite{gu2018nonautoregressive} on a larger dataset. (2) Our NACF surpasses NA-B by a large margin on both datasets. As we will show in the next subsection, this superior performance is attributed to the proposal of visual word generation, which not only generates informative gradients to promote the training of visual words but also alleviates the ``multi-modality'' problem by providing a warm start with the model during inference.

\subsection{Ablation Studies and Analyses}\label{sec:ablation_studies}
\textbf{Visual Word Generation.} Our proposed visual word generation task plays a critical role in both training and inference phases. During training, this task generates auxiliary gradients, thus the effect of $\mathcal{L}_{vis}$ on performance is worth exploring. As shown in Table \ref{tab:different_components}, $\mathcal{L}_{vis}$ brings promising performance gains for both NA-B (exp2 vs. exp1) and AR-B (exp5 vs. exp4), especially on the CIDEr-D metric. To figure out the word-level effect of $\mathcal{L}_{vis}$, first we collect the captions generated by the model with or without $\mathcal{L}_{vis}$, then we measure the relative growth rates of word frequency, and finally we average the results of words of the same type. As shown in Fig. \ref{fig:average_rgr}, visual words get an overall boost in various categories of videos. All these results demonstrate that training with $\mathcal{L}_{vis}$ can improve the caption quality by addressing the insufficient training of visual words.

During inference, visual word generation can produce coarse-grained ``templates'' (CT). An obvious improvement of taking CT as starting points can be observed in Table \ref{tab:different_components} (exp3 vs. exp2) and Table \ref{tab:different_decoding_algorithms}, which indicates that decoding with some known visual words can generate more semantically correct captions than the from-scratch generation that starts with completely unknown sequences. In summary, our proposed visual word generation is versatile and it benefits a lot for non-autoregressive video captioning. 

\begin{table}[t]
    \centering
    \begin{tabular}{ccccc}  
    \toprule
    Model &Novel &Unique &Vocab Usage\cr
    \midrule
    AR-B &17.19 &25.79 &3.36\cr
    NA-B &23.88 &31.24 &3.24\cr
    NACF &\textbf{34.35} &\textbf{42.47} &\textbf{3.83}\cr
    \bottomrule
    \end{tabular}
    \caption{Diversity of generated captions at various aspects (\%) on MSR-VTT.}
    \label{tab:diversity}
\end{table}

\begin{table}[t]
    \centering
    \begin{tabular}{ccccccc}  
    \toprule 
    Algorithm &B1 &B2 &B3 &B4 &M &C$_D$\cr
    \midrule

    MP  &81.0 &67.5 &53.5 &40.8 &28.2 &49.4 \cr
    EF  &81.5 &67.9 &54.1 &41.4 &28.7 &50.6 \cr
    L2R &81.0 &67.3 &53.4  &40.8 &28.4 &48.7 \cr
    
    \midrule
    CT-MP &\textbf{82.2} &\textbf{68.7} &\textbf{54.7} &\textbf{42.0} &28.7 &51.4 \cr
    CT-EF &82.1 &68.4 &54.4 &41.7 &\textbf{28.8} &\textbf{51.8}\cr
    CT-L2R &81.7 &68.2 &54.3 &41.7 &28.7 &50.6 \cr
    \bottomrule
    \end{tabular}
    \caption{Performance of our NACF using different decoding algorithms on MSR-VTT. Here $T = 5$ and $q = 1$.}
    \label{tab:different_decoding_algorithms}
\end{table}

\begin{figure*}[t]
\centering
\includegraphics*[width = 0.99\linewidth]{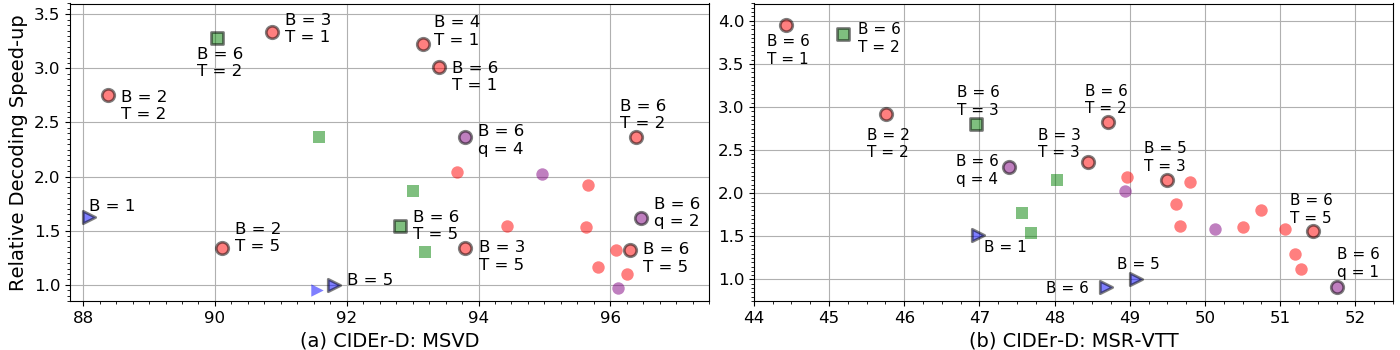}
\caption{Relative decoding speed-up versus CIDEr-D on (a) MSVD and (b) MSR-VTT. The circles in red denote NACF (CT-MP) while in purple denote NACF (CT-EF). The squares in green denote NA-B (MP). The triangles in blue denote AR-B. We take the latency of AR-B ($B=5$) as a benchmark, which costs 37.4 ms and 43.8 ms on MSVD and MSR-VTT, respectively.} 
\label{fig:speed_cider_trade_off}
\end{figure*}

\begin{figure*}[t]
\centering
\includegraphics*[width = 0.99\linewidth]{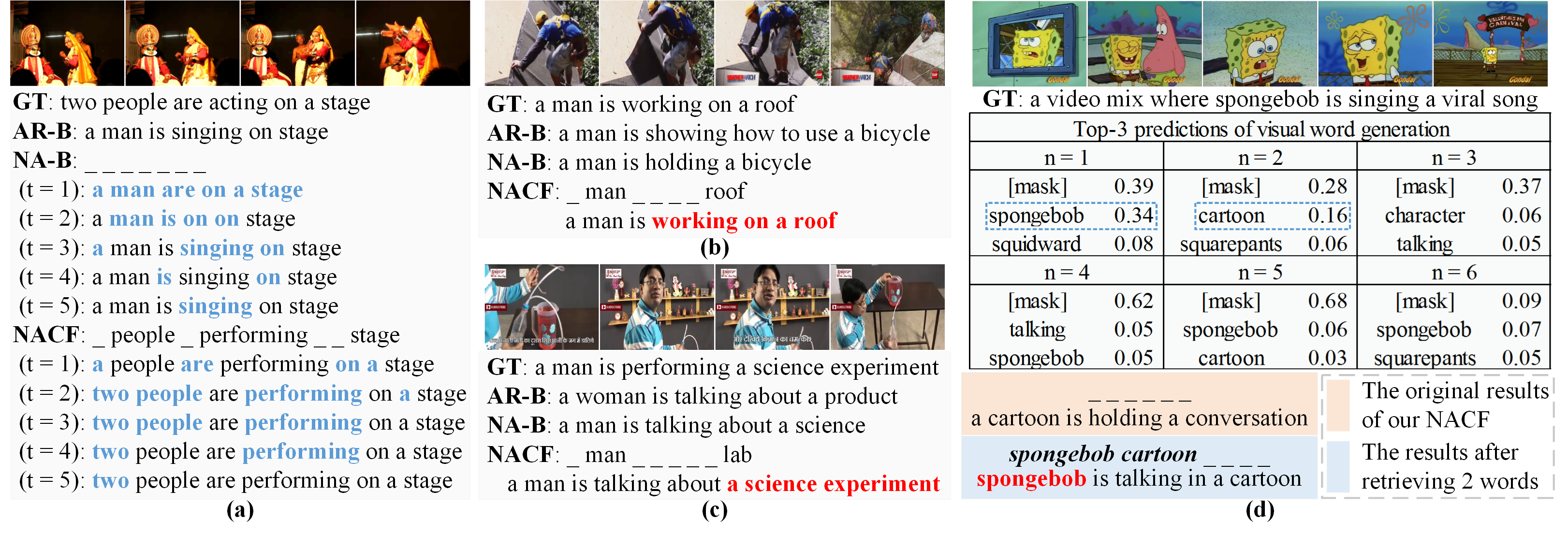}
\caption{Qualitative results on MSVD and MSR-VTT. The incomplete sentences of our NACF are generated coarse-grained ``templates'' (CT). In (a), words in blue denote the update during iterative refinement, while for the rest examples, accurate keywords are highlighted in red. In (d), our NACF generates an unsatisfying caption due to the inadequate generation of visual words, which can be alleviated by retrieving two visual words (``spongebob'' and ``cartoon'') that are potential to be predicted.} 
\label{fig:qualitative_results}
\end{figure*}

\textbf{Diversity.} To quantify caption diversity, we compute three metrics following \cite{dai2018neural}, namely \textit{Novel} (the percentage of captions that have not been seen in the training data), \textit{Unique} (the percentage of captions that are unique among the generated captions) and \textit{Vocab Usage} (the percentage of words in the vocabulary that are adopted to generate captions). As shown in Table \ref{tab:diversity}, NACF achieves the best performance across all metrics, indicating the advantage of our approach in terms of caption diversity.

\textbf{Decoding Algorithm.} In Table \ref{tab:different_decoding_algorithms}, we present the performance of all aforementioned decoding algorithms. As we can observe, (CT-)EF and (CT-)MP always outperform (CT-)L2R. Therefore, we can conclude that adaptive generation rather than monotonic generation is requisite for our NACF to generate plausible descriptions. 

\textbf{Inference Efficiency.} We measure latency\footnote{Latency is computed as the time to decode a single sentence without minibatching, averaged over the whole test set.} following \cite{guo2019non,wang2019non}, and conduct experiments in PyTorch on a single NVIDIA Titan X. Fig. \ref{fig:speed_cider_trade_off} shows the speed-performance trade-off on MSVD and MSR-VTT, where we take the latency of AR-B ($B=5$) as a benchmark. Notably, with CT-MP algorithm, our NACF can generate captions over 3.2 times ($B=4$, $T=1$) faster than AR-B ($B=5$) on MSVD while 2.2 times ($B=5$, $T=3$) on MSR-VTT without performance degradation. These results demonstrate the high inference speed of our NACF.

\subsection{Qualitative Analysis}\label{sec:qualitative_examples}
Fig. \ref{fig:qualitative_results} shows a few visualized examples of generated captions for different models. As we can see in (a), (b), and (c), while baselines AR-B and NA-B mistaken the video contents, our NACF can capture relevant visual words (\textit{e.g.}, ``roof'' in (b) and ``lab'' in (c)) and thus generate more precise captions. Specifically in (a), where the intermediate process of iterative refinement is presented, we can observe that the generated visual words of our NACF provide rich contextual information to predict the quantifier ``two''. However in (d), our NACF suffers from inadequate visual word generation. To figure out which visual words are potential to be predicted, we visualize top-3 predictions in (d) and find that most of them are somewhat relevant. According to the result that the description gets improved after retrieving ``spongebob'' and ``cartoon'', a more robust mechanism of generating visual words deserves further study.

\section{Conclusion}
In this paper, we propose a novel Non-Autoregressive Coarse-to-Fine (NACF) model for video captioning, which is based on a masked language model for parallelization and equipped with visual word generation and dedicated decoding algorithms to generate accurate and diverse captions in a coarse-to-fine manner. Extensive experiments on two video captioning benchmarks, MSVD and MSR-VTT, demonstrate that our proposed NACF achieves state-of-the-art performances with two unique advantages, \textit{i.e.}, generating more vivid descriptions and decoding faster (\textit{e.g.}, 3.2 times speed-up without performance degradation) than autoregressive captioning models. In our future study, we will focus on interpretable and controllable caption generation.

\section*{Acknowledgements}
Special acknowledgements are given to AOTO-PKUSZ Joint Research Center of Artificial Intelligence on Scene Cognition technology Innovation for its support.

\fontsize{9.8pt}{10.8pt} \selectfont
\bibliography{references}
\end{document}